# An Ensemble Blocking Scheme for Entity Resolution of Large and Sparse Datasets

Janani Balaji*, Faizan Javed*, Mayank Kejriwal**, Chris Min*, Sam Sander*, Ozgur Ozturk*
*Affiliation: CareerBuilderLLC, 5550 Peachtree Parkway, Norcross, GA 30092
**Affiliation: The University of Texas at Austin, Austin, TX 78712
*Email: {janani.balaji, faizan.javed, chris.min, sam.sander, ozgur.ozturk}@careerbuilder.com
**Email: mayankkejriwal@utexas.edu

## Abstract

Entity Resolution, also called record linkage or deduplication, refers to the process of identifying and merging duplicate versions of the same entity into a unified representation. The standard practice is to use a Rule based or Machine Learning based model that compares entity pairs and assigns a score to represent the pairs' Match/Non-Match status. However, performing an exhaustive pair-wise comparison on all pairs of records leads to quadratic matcher complexity and hence a Blocking step is performed before the Matching to group similar entities into smaller blocks that the matcher can then examine exhaustively. Several blocking schemes have been developed to efficiently and effectively block the input dataset into manageable groups. At CareerBuilder (CB), we perform deduplication on massive datasets of people profiles collected from disparate sources with varying informational content. We observed that, employing a single blocking technique did not cover the base for all possible scenarios due to the multi-faceted nature of our data sources. In this paper, we describe our ensemble approach to blocking that combines two different blocking techniques to leverage their respective strengths.

## 1 Introduction

In the online recruitment domain, recruiters source qualified candidates for their clients' job openings from a pool of both active and passive candidates. Active candidates are actively looking for work and hence they keep their personal brand updated. Passive candidates on the other hand are not likely to update their digital profile information, such as, resume and online social profiles, as frequently as active candidates, if at all. For proactive recruitment, recruiters usually leverage candidate search solutions from disparate vendors to identify qualified talent for their clients. Since each vendor may provide a different facet of a candidates' profile (e.g., social, technical, professional), recruiters usually want to aggregate such profile information to get a complete view of a candidate.

At CB, Recruitment Edge (Edge) is a one-stop search shop for recruiters that presents an enhanced view of a candidate by consolidating the information contained in a candidate's resume in our resume database, with their open web profiles. We aggregate candidate information and online activity from over 100 different sources and identify the ones that belong to the same candidate.

Entity Resolution (ER), also known as Deduplication or Record Linkage, is a long-established challenge in the Artificial Intelligence (AI) domain, where the goal is to provide accurate, fast and scalable solutions to identify and match duplicate entities. The core of the deduplication framework is the Matcher component, that functions as the logical decision unit to determine if the compared entities are a match or a non-match. The naive approach to deduplication involves the matcher performing a pair-wise comparison on all available entities, the results of which are aggregated together to represent duplicate instances. The matching process typically involves time-intensive syntactic/semantic data transformations and comparison operations and hence contributes a bulk majority of the total run time of an ER application. However, due to the massive volumes of data typically involved (we currently process about 0.5 billion profiles), conducting such exhaustive comparisons is not a scalable option. Our business needs also necessitate a faster turnaround time to avoid presenting users with stale data.

To tackle this, traditionally, a Blocking process is applied, where, seemingly similar profiles are grouped together into smaller sized blocks, which are then exhaustively explored by a Matcher component. So, instead of performing ($n^2$) comparisons, the matcher only performs ($p.m^2$) comparisons, where p is the number of blocks and m $<<$ n. The Blocker, forms a key component in the ER pipeline, as it determines the entities that would be examined by the matcher, hence directly influencing the efficiency and effectiveness of the entire ER process.

Several blocking schemes have already been proposed and widely used to reduce matching complexity, without impacting the recall of the ER framework. However, in our application, since we procure data not only from multiple sources, but also with each source representing different aspects of a person (Social, Professional, Technical), employing a single blocking method proved insufficient to guarantee comprehensive duplicate coverage. We experimented with two popular blocking techniques and found that, us- ing only either one of them, resulted in loss of a few poten- tial duplicate pairs from being fed into the Matcher stage, thereby reducing the effectiveness of the entire deduplication process. While one blocking method was well suited for sparse, heterogeneous datasets, it required apriori knowl- edge of the relative importance of attributes. Also, the run time complexity prevented us from employing it extensively. The other technique, though less complex and generic, does not guarantee complete pair coverage for large datasets. In this paper, we propose an ensemble blocking method by leveraging two different blocking techniques to produce a robust blocking framework that is resistant to data heterogeneity.

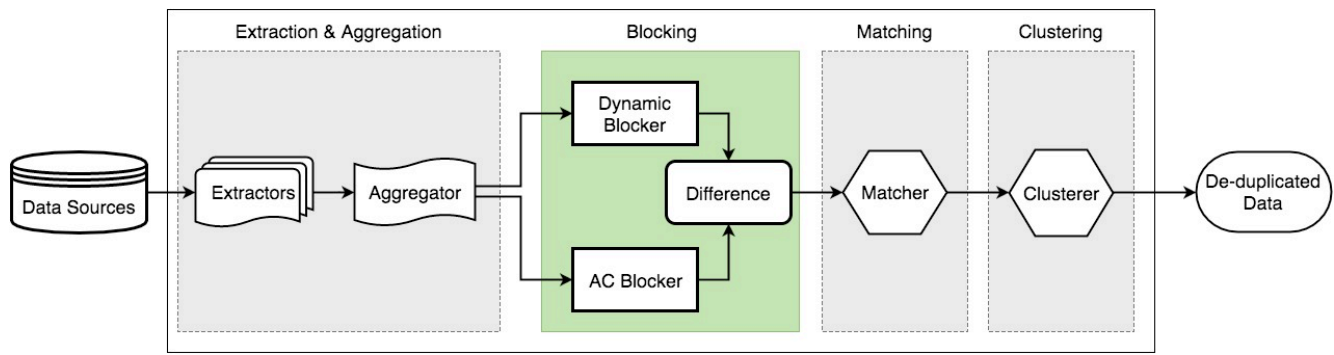

Figure 1: A high level schematic view of the ER pipeline

Section 2 briefs over the current state of art of the blocking algorithms, Section 3 describes the ER process employed at CB and establishes the need for our ensemble blocking approach. The experiments performed on real and synthetic data, given in Section 4, gives insight into the benefits observed from the proposed approach and Section 5 concludes the paper with directions for further research.

## 2 Related Work

Surveys of well-known blocking techniques such as sorted neighborhood and canopy clustering (among others) can be found in (Christen 2012) (Elmagarmid, Ipeirotis, and Verykios 2007). The sorted neighborhood approach assumes that duplicate records will be close to each other in a sorted list. In this approach, records are first sorted by an individuating blocking key. A sliding window of a fixed size is then sequentially moved through the sorted records and matching is limited to records that fall within the window. In canopy clustering, records are first grouped into overlapping blocks using a computationally cheap similarity comparison metric. This is followed by applying a more expensive similarity metric for pairwise comparisons within blocks. More recently, there has been a trend towards incorporating crowdsourcing frameworks in entity resolution (ER) pipelines. Many such human-powered ER approaches do not incorporate a blocking step. However, human-powered blocking can result in lower crowdsourcing costs while maintaining high accuracy (Li, Lee, and Lee 2014). Corleone (Gokhale et al. 2014) leverages human-powered blocking via crowdsourced active learning to learn blocking rules.

Attribute Clustering (AC) blocking (Papadakis et al. 2013) is an unsupervised technique that addresses the issue of efficiency vs effectiveness in large datasets. It tokenizes the entire data into a schema-agnostic bag of tokens format. All entities that share the same token are then grouped into the same block. This schema-independent technique offers a straightforward, scalable method for blocking massive volumes of data.

Dynamic blocking (McNeill, Kardes, and Borthwick 2012) is another alternative that works especially well with sparse, heterogeneous datasets. It works by iteratively splitting large blocks into smaller ones by adding additional attributes to the blocking key to produce blocks with reasonable uniqueness. This process of creating sub blocks from larger blocks continues until the block size is below a predefined threshold. However, this technique is inherently iterative in nature and becomes time-intensive for large datasets with larger number of attributes.

While AC and Dynamic Blocking are both effective blocking schemes in their own right, due to the vast difference in the characteristics of our data sources, we observed that neither was able to guarantee comprehensive duplicate coverage for all kinds of data. In this paper, we describe our ensemble blocking approach that combines our in-house variants of the AC blocker and the Dynamic blocker to effectively produce a blocking methodology that performs consistently well for different datasets. Our approach is similar to (Papadakis et al. 2013) in that we define a blocking framework which leverages multiple blocking algorithms.

## 3 System Overview

The Entity Resolution process at CB follows the abstract schematic given in Figure 1. The Extraction and Aggregation phases collect, de-noise and normalize the input records into a common representation that can be used in the subsequent stages. The second phase is the Blocking process, which groups records into manageable blocks based on shared attributes. The Matcher performs an exhaustive pair-wise comparison on each block and classifies each profile pair, into a Match or a Non-match using a set of heuristic rules. The Clusterer aggregates the individual match pairs into clusters representing the same entity. This paper elaborates on the second step in the pipeline - the Blocker. The other stages of the pipeline are not covered here.

The aim of the blocking step is to ensure that all potential duplicate entities are grouped together in at least one block, so that the matcher can have a shot at evaluating them. However, a blocker that groups all available entities into one massive block, might guarantee complete coverage, but fails in reducing matcher complexity. Similarly, a highly restrictive blocker might dramatically reduce the number of comparisons, but has a higher risk of losing potential duplicate pairs. Finding the right balance between the two is not a trivial task. This issue is further compounded due to the heterogeneity and sheer volume of our data.

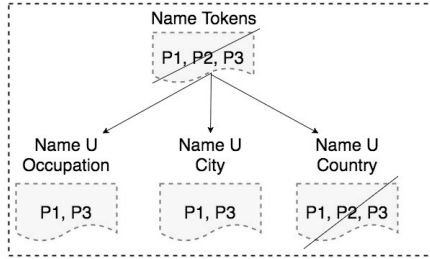

Figure 2: Blocking Schemes: A Comparison

We procure our data from a large number of heterogeneous sources with varying data collection practices, resulting in inherently sparse and noisy data. While that is quite common in large scale ER applications, we have an additional challenge in the form of the disparate nature of the content that each of these sources offer. In order to obtain comprehensive unification, we gather data from sources that offer information about different facets (Social, Professional, Technical, etc.,.) of a person. Hence, our input user profiles show large variances in attributes and are characterized by omissions, misspellings and incomplete/stale data. For example, a person might not choose to provide his entire employment history on his Social profile. He might also go by his full legal name on his Professional profile, while choosing to provide his nick name in his Social/Technical profiles. Due to such high variance in informational content, it becomes impossible to define a minimal set of attributes that are consistent among profiles from all facets of interest, based on which deduplication could be done. This introduces additional complexities in determining a generalized blocking key that could result in uniform sized blocks without compromising on pair completeness.

AC blocking offers a straightforward, scalable blocking scheme that is easily adoptable in distributed frameworks like Spark and Mapreduce. The original method called for a schema-agnostic block creation to account for heterogeneous data. Since our input sources are schema-normalized, we modified the method to include schema definitions. We created unique blocking keys by attaching the attribute information along with the value.

For example, consider the database shown in Figure 2(a) that lists five different profiles along with their attributes. Figure 2(b) shows the blocks created by our version of the AC method, while the original AC would have created blocks shown in Figure 2(c). For brevity, only those blocks with at least two entities are shown in the figure. While the AC blocking method (Papadakis et al. 2013) uses a bag-of-tokens approach to tokenize the entities, in doing so, the blocks are often proliferated with spurious pairs. Since our extraction and aggregation phases are well-defined, we take advantage of the apriori schema knowledge to further denoise our blocks. As we see, including the schema information in block creation eliminated the erroneous block highlighted in red that was created by the original AC method.

However, for all its simplicity, the resulting block sizes from AC can be quite large for blocks based on common-valued attributes. In practice, those blocks with size above a given threshold are discarded through block purging. The idea behind discarding larger blocks is that they offer little informational value and are more likely to hold redundant duplicates, i.e., the large blocks will not contain entity pairs that are also not present in a smaller block. However, for data from heterogeneous sources, this assumption does not necessarily hold true. When information from varied data sources representing different facets is aggregated, it is often the case that each source contains information only on a subset of attributes. Due to the large number of omissions and variances in attribute values, discerning information that can place two potential duplicates in the same block might be absent, resulting in losing potential matching pairs.

For example, while a person's professional profile might provide complete information about his employment history, his social profile need not. So, the only information that can group these two profiles within the same block

Table 1: Performance Comparison

| Datasets | | | AC Blocker | | | Dynamic Blocker | | | Ensemble Blocker | | |
|---|---|---|---|---|---|---|---|---|---|---|---|
| Name | Size | Type | PC | RR | Time | PC | RR | Time | PC | RR | Time |
| Febrl-1 | 2000 | S | 1.0000 | 0.9923 | 5153 | 0.9834 | 0.9962 | 10327 | 1.0000 | 0.9904 | 11302 |
| Febrl-2 | 3000 | S | 1.0000 | 0.9957 | 4163 | 0.9694 | 1.0000 | 12713 | 1.0000 | 0.9945 | 17143 |
| Febrl-3 | 4000 | S | 1.0000 | 0.9967 | 4150 | 0.9674 | 0.9968 | 16785 | 1.0000 | 0.9959 | 21019 |
| Dataset-1 | 258 | R | 1.0000 | 0.8873 | 438 | 0.7310 | 0.9967 | 720 | 1.0000 | 0.8873 | 1158 |
| Dataset-2 | 299 | R | 1.0000 | 0.9907 | 1503 | 1.0000 | 0.9993 | 3000 | 1.0000 | 0.9905 | 4503 |
| Dataset-3 | 396 | R | 1.0000 | 0.9198 | 648 | 1.0000 | 0.9972 | 1158 | 1.0000 | 0.9198 | 1806 |
| Dataset-4 | 1468 | R | 1.0000 | 0.9779 | 3484 | 0.8963 | 0.9995 | 10207 | 1.0000 | 0.9778 | 13691 |
| Dataset-5 | 1991 | R | 0.9833 | 0.9943 | 4200 | 0.9966 | 0.9981 | 11610 | 1.0000 | 0.9929 | 15810 |
| Dataset-6 | 5892 | R | 0.8909 | 0.9990 | 19622 | 1.0000 | 0.9998 | 38509 | 1.0000 | 0.9989 | 58131 |

Metrics and Symbols: Pair Completeness (PC), Reduction Ratio (RR), Time (ms), Synthetic (R), Real (R)

might be generic enough to create a large block which eventually gets purged. For example, in Figure 2(b), the block Country:UnitedStates is based on a highly common key and hence places all the profiles into the same block. If we set the block threshold to be 2, then the only block retained is the one based on Occupation:Teacher. However, this causes us to lose a valid duplicate pair (P1,P3), which is caught only in block Country:UnitedStates . This is because, P3 is missing individuating information in both its Occupa- tion and City data and can be matched with P1 only based on Country:UnitedStates. However, Country:UnitedStates is too generic, that it loses its significance as a stand-alone blocking key. We have observed such omissions to be the norm instead of a rarity while aggregating data from different sources. Hence, in spite of the scalability of the AC blocker, due to our multi-faceted data sources, we were unable to use it as the sole blocking technique.

The Dynamic blocker is especially well-suited for large, heterogeneous data and seemed like a good fit for our framework. It groups profiles into blocks based on a hierarchical organization of attributes. We used name as the top-level blocking criteria and further broke down the larger blocks by adding additional individuating attributes as blocking keys. This solved the problem of sparse information getting lost in large blocks by defining more expressive blocking keys to create blocks with reasonable uniqueness. However, the number of blocking keys can grow exponentially with the number of attributes and their values, leading to a sharp increase in runtime with the addition of additional attributes. This caused us to restrict the number of attributes that could be used for blocking, thereby failing to leverage the entire range of our data. Also, due to the hierarchical nature of the process, it required apriori knowledge about the relative importance of the attributes, as the choice of the primary attributes at each hierarchical level determines the effectiveness of the entire blocking algorithm. The heterogeneity of our data sources once again negatively impacts this situation due to inconsistencies in the kind of attributes. Dynamic blocking, therefore, is highly susceptible to the presence of noise, especially in the higher levels of hierarchy.

For example, in Figure 2(a), P4 and P5 belong to the same person. However, both these profiles have different values in the First Name and Last Name fields. This is a highly common scenario, especially with female profiles, due to surname changes. The presence of nick names also complicates this issue. As shown in Figure 2(d), since the name attributes do not match and since name is selected to be the first blocking criterion, the two profiles end up not being placed in the same block. This results in a potential duplicate pair that is lost at the early stages of the process. Real world data is riddled with such errors and missing such matches could cause a serious ramifications in the quality of the deduplication process. The AC method described above would not have missed this pair, as it would have grouped them based on the common attribute occupation:Teacher, as shown in Figure 2(b).

### 3.1 Ensemble Blocking Approach

While the Dynamic blocker handles problems due to sparseness, the complexity involved prevents us from employing more attributes for the blocking stage. Also, the hierarchical nature causes inflexibility in dealing with common data deficiencies. The AC blocker on the other hand is efficient for large datasets with more number of individuating attributes. However, it misses out from grouping a few potential duplicate pairs due to absence of discriminating information. We could extend the impact of the Dynamic blocking by increasing the range of attributes used at each hierarchy. However, this results in a steep increase in our run time. To combat these issue, we use a combination approach by combining the results of both the AC blocker as well as the Dynamic blocker.

Since the Dynamic blocker is better suited at handling large sparse data, which is more characteristic of our datasets, we use it as our primary blocking technique and use the AC blocker as a catch-all net for those attributes that are not handled by the Dynamic blocker. This lets us designate a subset of the most discerning attributes to be used in the Dynamic blocker and use the remaining in the AC blocker to catch those pairs potentially missed by the Dynamic blocker. The ensemble approach is shown in Figure 1. The cleaned and normalized data is fed to both the Dynamic as well as the AC blockers. The entire set of candidate pairs generated by the Dynamic blocker are fed into the subsequent matcher, while only the additional pairs generated by the AC blocker that were not generated by the Dynamic blocker go through the matching process.

The Dynamic blocker only uses the name features as the top-level blocking attributes and creates sub blocks by combining name with each of the other attributes used. We used 13 attributes for the Dynamic blocker. The AC blocker on the other hand uses all the available 17 attributes, thereby utilizing additional information not used by the Dynamic blocker due to run time concerns. Since there is a large variation in our data sources, employing only one of the above blocking techniques would cause us to lose potentially duplicate pairs. Using an ensemble approach, on the other hand, lets us cover the deficiencies of both the blocking techniques.

Our experiments show that this combination approach helps increase blocking effectiveness without taking a deep hit on efficiency. The details of our matcher phase, which labels each pair into a match or a non-match, are not covered in this paper.

## 4 Experiments

We experimented with using the Dynamic and AC blockers independently and also with using the ensemble approach by combining the results of both the blocking techniques. In this section, we share our findings and the improvements observed in using the combination approach.

### 4.1 Measures

Traditionally, blocking techniques have been evaluated on effectiveness and efficiency. Effectiveness is defined by the total coverage of duplicate pairs produced by the blocker. The Pair Completeness (PC) metric, defined as $\frac{|D_B|}{|D_G|}$, where $|D_B|$ represents the number of pairs produced by the blocker and $D_G$ represents the number of pairs in the gold standard, severs as an indicator of the effectiveness of the blocking technique. The PC metric takes values in the range [0,1], with a value closer to 1 indicating better coverage. Measuring the efficiency gives an indication of how well the blocker performs its primary task - to reduce the complexity in the matcher phase by generating a small candidate set that is a subset of the exhaustive set of all input candidate pairs. The Reduction Ratio (RR), defined as $1 - \frac{|D_B|}{|D_{ALL}|}$, where $D_B$ represents the number of pairs produced by the blocker and $D_{ALL}$ is the cardinality of the set of all combinations of the input candidate pairs, provides insight into the effectiveness of the blocker. We use these two metrics to illustrate the performance of our blocking scheme.

### 4.2 Datasets

We used 3 synthetic datasets and 6 manually labeled datasets to illustrate the advantages of our ensemble approach. The synthetic datasets (Febrl 1-3) were generated by the Febrl (Christen 2008) tool. The manual datasets (Dataset 1-6) were randomly sampled from our database and manually labeled to identify duplicate profiles. Both the datasets had 17 attributes including First Name, Last Name, City, Employer and Job Title among others.

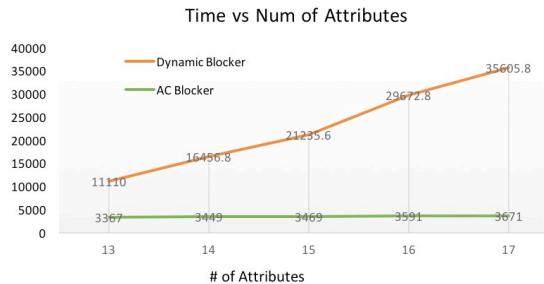

Figure 3: Scalability with Num of Attributes

### 4.3 Discussion

The results of our experiments are given in Table 1. These experiments were performed by using all the 17 attributes for the AC blocker and 13 attributes for the Dynamic Blocker. The name tokens were used as the top level blocking criterion for the Dynamic Blocker. We noticed an interesting trend in the blocker performance for the different datasets. The AC blocker consistently seems to perform better than the Dynamic blocker in PC for all datasets except for Dataset 5 and Dataset 6. These are the two largest datasets in our collection and therefore have higher probabilities of forming larger blocks which eventually get discarded. The Dynamic blocker shows better PC in such cases. However, the AC blocker has a more flexible framework that is robust to most data deficiencies. This is reflected in its consistently good PC metric for the majority of the datasets. While the AC blocker has good performance for smaller datasets, the PC metric starts to steadily decline with increasing data volumes. Since real world data is unquestionably large, the AC blocker alone would not guarantee complete pair coverage.

Comparing the RR metric in Table 1, we see that the Dynamic blocker has a higher RR for all scenarios. However, the reduced PC metric denotes that increase in RR for the Dynamic blocker comes at the cost of missing out a few potentially duplicate pairs. While both PC and RR are important metrics to evaluate blocker performance, for business reasons, we lean more towards having better PC, as long as the reduction in RR is not too huge.

Table 1 also compares the Ensemble blocker we described in Section 3.1. We see that using both the blockers gives complete pair coverage for all the datasets (PC=1). Though we do see a decrease in RR compared to the Dynamic blocker, the reduction is not too significant to cause substantial difference in matcher complexity. It also proves our theory that the Dynamic blocker and the AC blocker together can account for most real world scenarios in our ER pipeline. Table 1 also lists the running time for all three blocking methods. We see that the total running time for the ensemble method is comparable to the Dynamic Blocker, with the AC blocker taking the shortest run time. On average, we see that the AC blocker is 62% faster than the Dynamic blocker. Note that the Dynamic blocker worked with 13 attributes, while the AC blocker operated on 17 attributes.

We also studied the scalability for the blocking techniques by varying the number of attributes used for blocking. Figure

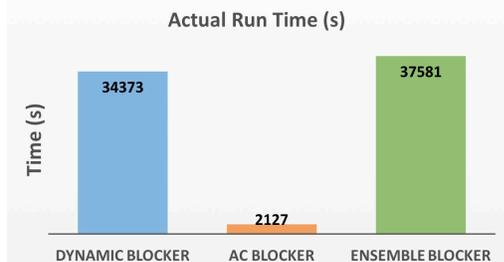

Figure 4: Distributed Run Time

3 shows the running times for the Dynamic and AC blockers for different attribute sets on Dataset-6, our largest dataset. The time reported is the average over 10 runs. While the Dynamic blocker shows a linear increase in runtime with addition of more attributes, the AC blocker showed less sensitivity to attribute size. This explains our reasoning for fixing the attribute size of the Dynamic blocker at 13 and using the full set of attributes for the AC blocker.

### 4.4 Distributed Run Times

The experiments discussed above were carried out in a laptop running on a 3.1 GHz Intel Core i7 processor with 16GB of ram. While a single processor environment is sufficient to illustrate the general trends of the two algorithms for the test datasets, it is not a viable option for our entire data pool. If we were to extrapolate the running times for the test datasets into our actual dataset (~ 0.5B), it would theoretically take months to block our entire data. We use the Spark distributed framework [1] to run our entire ER pipeline, and Figure 4 shows the run times for the individual blockers and our combination approach in a distributed environment.

As seen, the AC blocker is 93% faster than the Dynamic blocker, though it operates on more number of attributes(13 for Dynamic and 17 for AC). This difference is much greater than the 62% observer earlier for the single processor case. This is because, the blocking scheme of the AC blocker is inherently more tuned towards faster distributed implementations in the Map-Reduce format due to lack of dependency between blocking attributes. The Dynamic blocker, on the other hand, is iterative in nature, with each round of blocking dependent on the previous blocking result. This makes it difficult to scale well in a distributed environment, leading to increased run time.

The run times for the ensemble blocker, on the other hand, is comparable with the run time of the Dynamic blocker, though it utilizes the complete set of attributes, in the form of the AC blocker. While using only the AC blocker would have drastically cut down our run times, it comes with the cost of a reduced PC for large data volumes as observed in our test datasets.

To further give insight into the need for the ensemble approach, we observed that using both the Dynamic and AC blockers increased the number of de-duplicated pairs by 30%, compared to using only the Dynamic blocker. This is a significant increase, which has a direct impact on the recall rate of our ER process, by increasing the total number of de-duplicated entities.

## 5 Conclusion

In this paper, we described different blocking techniques currently being used in the ER pipeline at our organization and established that using a single blocking scheme was not able to guarantee sustained pair coverage for all kinds of data. We demonstrated that using an ensemble approach by combining the Dynamic and AC blockers produces a more robust blocking framework that is less susceptible to data variations. Currently, except for the name features in Dynamic blocking, all our blocking keys enjoy equal weightage in block determination. In future, we would like to study the individual impact that each attribute has towards blocking quality and devise intelligent key-combinations that work well for heterogeneous data types.

---

[1] Zaharia, M.; Chowdhury, M.; Franklin, M. J.; Shenker, S.;and Stoica, I. 2010. Spark: Cluster computing with working sets. In Proceedings of the 2Nd USENIX Conference on Hot Topics in Cloud Computing, HotCloud10, 1010. Berkeley,CA, USA